\documentclass[letterpaper, 10 pt, conference]{ieeeconf}  

\usepackage{amssymb}
\usepackage{amsmath}
\usepackage{color}
\usepackage{graphicx}
\usepackage{cite}
\usepackage{float}
\usepackage[ruled, linesnumbered]{algorithm2e}
\usepackage{booktabs}

\usepackage{amsmath}
\usepackage{bm}
\usepackage{url}

\graphicspath{ {./images/} }

\IEEEoverridecommandlockouts                              
\title{\LARGE \bf
Redefining Data Pairing for Motion Retargeting \\ Leveraging a Human Body Prior
}

\author{Xiyana Figuera$^{*1}$, Soogeun Park$^{*2}$, and Hyemin Ahn$^{2}$
\thanks{$^{1}$Xiyana Figuera is with the Department of Computer Science and Engineering, Ulsan National Institute of Science and Technology, Ulsan, Korea. {\tt\small{ xyin@unist.ac.kr}}.}%
\thanks{$^{2}$Soogeun Park and Hyemin Ahn are with Graduate School of Artificial Intelligence, Ulsan National Institute of Science and Technology, Ulsan, Korea. {\tt\small{ \{soogeun, hyemin.ahn\}@unist.ac.kr}}.}%
\thanks{*Equal contribution}%
}

\begin{document}

\maketitle
\thispagestyle{empty}
\pagestyle{empty}


\begin{abstract}
We propose MR.HuBo (Motion Retargeting leveraging a HUman BOdy prior), a cost-effective and convenient method to collect high-quality upper body paired $\langle \text{robot, human} \rangle$ pose data, which is essential for data-driven motion retargeting methods. 
Unlike existing approaches which collect $\langle \text{robot, human} \rangle$ pose data by converting human MoCap poses into robot poses, our method goes in reverse. 
We first sample diverse random robot poses, and then convert them into human poses. However, since random robot poses can result in extreme and infeasible human poses, we propose an additional technique to sort out extreme poses by exploiting a human body prior trained from a large amount of human pose data. Our data collection method can be used for any humanoid robots, if one designs or optimizes the system's hyperparameters which include a size scale factor and the joint angle ranges for sampling. In addition to this data collection method, we also present a two-stage motion retargeting neural network that can be trained via supervised learning on a large amount of paired data. Compared to other learning-based methods trained via unsupervised learning, we found that our deep neural network trained with ample high-quality paired data achieved notable performance. Our experiments also show that our data filtering method yields better retargeting results than training the model with raw and noisy data.
Our code and video results are available on \textcolor{blue}{\url{https://sites.google.com/view/mr-hubo/}}.
\end{abstract}

\section{INTRODUCTION}
Motion retargeting, a fundamental task in computer animation and robotics, aims at finding a mapping between motions across different domains. 
This field has garnered considerable interest due to its wide-ranging applications, including avatar control in virtual environments, motion-related computer graphics tasks, and tele-operation in robotics.

With the pervasive integration of deep learning (DL) into robotics and computer vision, a myriad of studies have addressed this task employing DL methodologies \cite{villegas2018neural, lim2019pmnet, choi2020nonparametric, choi2021self, zhu2022mocanet}.
However, the application of these in motion retargeting for robotics, has been largely restricted to employing semi-supervised or unsupervised learning techniques. The current restriction stems from the challenges inherent in collecting paired $\langle \text{robot, human} \rangle$ pose data. 
\begin{figure}[thpb]
    \centering
    \hspace*{-0.1cm} 
    \raisebox{-\height}{\includegraphics[height=0.295\textwidth]
    {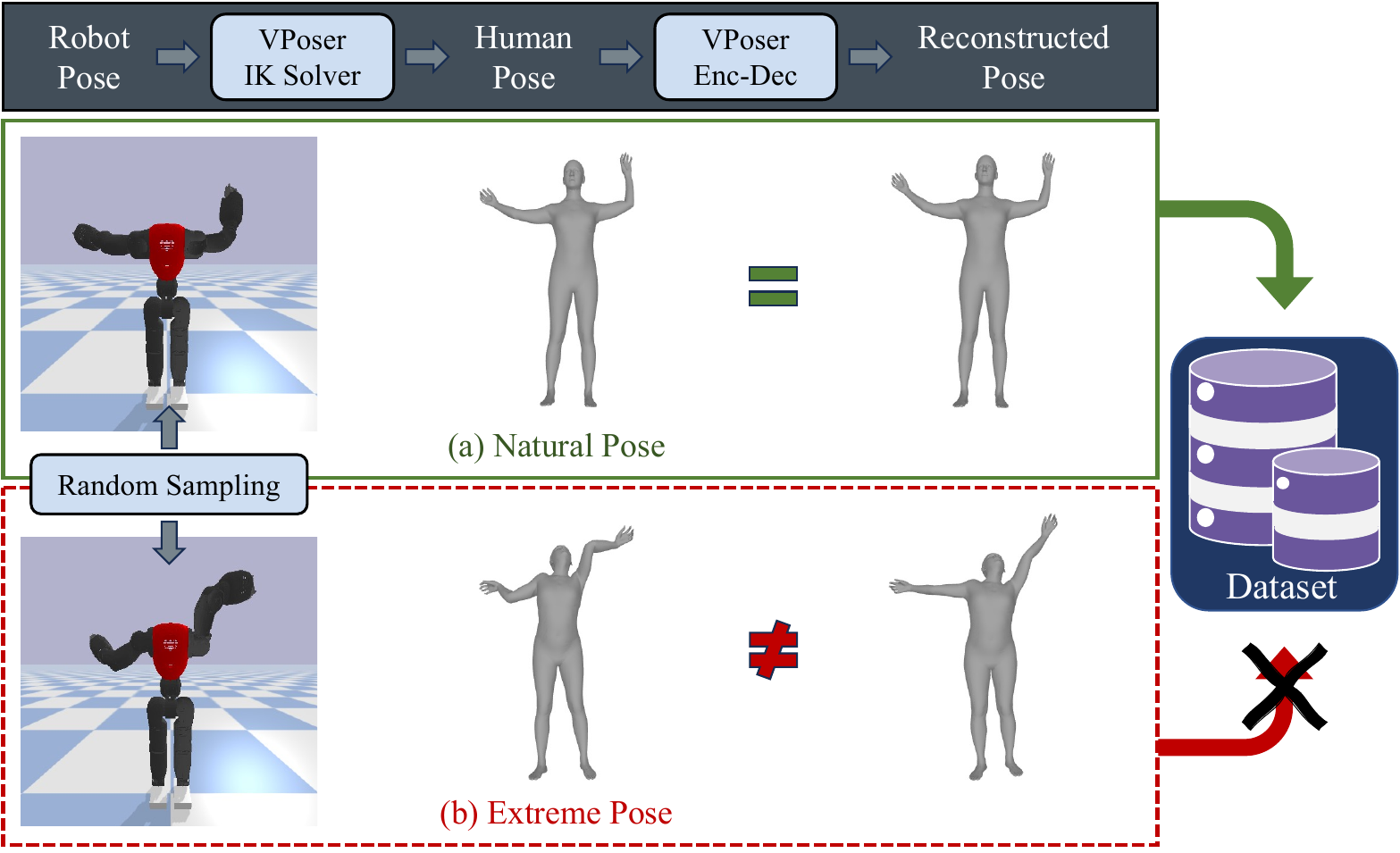}} 
    \caption{Visualization of how our data sampling method works. It randomly samples a robot pose (left) and converts it to the respective human pose (middle). Then, the obtained pose is reconstructed to a pose without noise (right). When random robot pose sampling results in extreme poses (below), we found the difference between the original and denoised poses becomes larger. Based on this, we collect only feasible poses (top) whose differences with denoised poses are small.
    }
    \label{fig:1_Extreme_poses_example}
    \vspace{-4mm}
\end{figure}
There exist three main hindrances in paired dataset collection: (1) the difficulty in defining ground truth robot motion for input human motion, (2) the limitation in the variety of poses that arises from the dependence on human motion data, and (3) the necessity to label this data for each robot hardware platform, as robots possess unique joint configuration space. 

\textcolor{black}{Recognizing these obstacles, our work centers on enabling computationally efficient supervised learning in motion retargeting, by \textcolor{black}{proposing} an automatic method for paired $\langle \text{robot, human}\rangle$ pose data collection. We \textcolor{black}{implement} a paradigm shift similar to the one in \cite{choi2021self}, wherein randomly sampled robot poses are converted into human poses with strategies to avoid extreme poses. Our method is versatile, and applicable to any humanoid robot, as long as the hyperparameters related to kinematic chain, link lengths, and size scale factor can be manually designed or optimized.}

To be more specific, we randomly sample robot poses from the valid joint angle range and convert them into human poses. However, this randomness can result in a human pose that is physically implausible or extreme (see Fig \ref{fig:1_Extreme_poses_example}-(b)), \textcolor{black}{which} can hamper the training of the motion retargeting model by introducing noise. To compensate for this, we propose a strategy to obtain natural and feasible human poses with minimal or even no reliance on glue data (i.e., paired samples obtained from classic optimization-based methods), which \cite{choi2021self} necessarily requires. 
To achieve this, we employ a human body prior (VPoser) \cite{SMPL-X:2019} which is a variational autoencoder that has learned a prior of SMPL \cite{SMPL:2015} human body pose parameters from massive human pose data \cite{mahmood2019amass}. 
We found out VPoser can be used (1) as an inverse kinematics (IK) solver which can find SMPL pose parameters from robot joints' XYZ position, and (2) as a denoising autoencoder, which converts implausible and noisy poses into valid ones.

Figure \ref{fig:1_Extreme_poses_example} shows how our data pairing method works in short. 
First, we randomly sample robot joint angles from their valid range (left of Fig. \ref{fig:1_Extreme_poses_example}). 
Then, we use VPoser as an inverse kinematics solver to obtain corresponding human pose (center of Fig. \ref{fig:1_Extreme_poses_example}). 
Here, our concern is that the human pose can become infeasible (Fig. \ref{fig:1_Extreme_poses_example}-(b)). 
Therefore, to sort out infeasible poses, we employ VPoser as a denoising autoencoder and obtain the reconstructed or denoised human pose. 
If the distance between the original and denoised human poses is large, we do not add it to our dataset since it indicates the generated pose is unusual.

After collecting paired data, we introduce a supervised learning-based motion retargeting network trained with a dataset obtained with our proposed data pairing method. This network employs a two-stage neural architecture that exploits 6D representations \cite{zhou2019continuity}, leveraging the capacity of neural networks to learn more effectively from continuous representations.
Additionally, our method can be used for motion retargeting in real-time with RGB cameras, when combined with a regression-based human mesh recovery network \cite{pymafx2023} which can extract SMPL pose parameters from RGB images.

Our experimental results show that our proposed method demonstrates a decent performance compared to an unsupervised method baseline \cite{choi2021self} and even surpasses it when measuring the performance based on the prediction error in link's XYZ positions. 
In addition to that, our ablation studies indicate that removing noisy poses helps improve the performance of the model, especially when the number of joints and size of the limbs of the robot kinematics differs significantly from that of a human.

Our contributions can be summarized as follows:
\begin{enumerate} 
\item We propose a data pairing method that can be applied to diverse humanoids with different kinematics and topology and can be used to efficiently obtain a large amount of high-quality $\langle \text{robot, human} \rangle$ data. %

\item We enable a deep learning-based motion retargeting model to learn in a fully supervised, end-to-end way with the use of a large amount of paired data.

\item  We propose a two-stage network that harnesses a continuous representation of 3D rotations (i.e. 6D representation) to describe human and robot poses, facilitating the learning of the network.

\end{enumerate}

\section{RELATED WORK}
\subsection{Optimization-based Motion Retargeting}

A wide range of existing literature addresses the task of motion retargeting by defining it as an optimization problem with constraints such as joint limits, usually to solve an inverse kinematics (IK) problem \cite{dariush2008online, suleiman2008human}. In this section, inspired by \cite{wang2019generative}, we categorize the optimization-based motion retargeting literature, by the core system used to capture human motion, into Marker-systems, Sensor-systems, and Camera-systems.

\textbf{Marker-systems}. Attaching markers to the human body was a commonly used system to capture human motion in motion retargeting literature. For instance, Riley et al. \cite{riley2003enabling} achieved real-time motion retargeting facilitated by simple visual markers (i.e. color patches) placed on the upper body. They tracked movements with 3D vision and incorporated a human kinematics model, solving the motion retargeting task through IK. Physical markers attached to the body were also used by Ott et al. \cite{ott2008motion}. In their work, the authors proposed a Cartesian control approach by substituting IK with a Hidden Markov Model (HMM), and optimizing motion primitives—parameters of the HMM—via expectation maximization for controlling the upper body joints. For lower body joints, a center of gravity-based balancing controller via optimization was proposed.

\textbf{Sensor-systems} 
Recent works employ modern devices such as Xsens MVN, which is a motion capture system that captures human motion through inertial and magnetic sensors \cite{roetenberg2009xsens,schepers2018xsens}. Koenemann et al. \cite{koenemann2014real} proposed a system that employs a compact human model obtained from such modern devices. They solved IK while considering the center of mass of the human to find statically stable configurations. To enhance the scalability of motion retargeting methods (i.e. retargeting motion to different robots from different human subjects), Darvish et al. \cite{darvish2019whole} proposed a method for whole-body retargeting that involves mapping anthropomorphic motions of human links to corresponding robot links through IK, ensuring the preservation of geometric relationships between them.

\textbf{Camera-systems} Many types of cameras have been adopted for motion retargeting, such as time of flight (ToF) imaging devices \cite{dariush2008online}, stereo cameras \cite{lim2022online} and RGB cameras \cite{khalil2021human}. 
Dariush et al. \cite{dariush2008online} addressed the task of online motion retargeting via a Cartesian space control theoretic approach with the use of low-dimensional human task descriptors from depth image sequences. Focusing on pose estimation from uncalibrated videos, Khalil et al. \cite{khalil2021human} proposed a motion retargeting method with input from a single-view RGB camera. The method consists of three modules that are designed for 2D coordinate extraction, depth estimation, and human joint angle computation. Yet, this method is limited to humanoid robots that have a close number of DoFs to those of humans.

Optimization-based retargeting approaches are also being used to collect $\langle \text{robot, human}\rangle$ pose paired data for data-driven methods by harnessing human MoCap data \cite{choi2020nonparametric,choi2021self}. 
However, optimization-based methods are usually not computationally efficient \cite{zhang2022kinematic}. 
Our data pairing method, though optimization-based, overcomes this limitation by employing VPoser \cite{SMPL-X:2019} which optimizes a low dimensional latent vector instead of full SMPL pose parameters when obtaining the paired dataset. 

While our retargeting method is not optimization-based, we opted for a camera system among the various systems in the aforementioned works. This choice allows us to harness a regression-based human mesh recovery network \cite{pymafx2023} at the inference stage whose output is SMPL pose parameters.

\subsection{Data-driven Motion Retargeting}
In recent years, data-driven motion retargeting has emerged as a prominent research area not only in robotics but also in computer vision, intending to achieve natural and scalable motion adaptation across various source and target subjects\cite{choi2021self,wang2019generative}. 
One of the advantages of data-driven methods is that, at the execution phase (i.e. inference phase), the computation can be forward-propagated in a single time which leads to faster computation than iterative IK-based methods \cite{choi2021self}. 
However, the expensive costs of collecting $\langle \text{robot, human}\rangle$ paired datasets have limited the research to employing unsupervised or semi-supervised methods \cite{villegas2018neural,lim2019pmnet}. 
To circumvent the need for paired data, Villegas et al. \cite{villegas2018neural}
proposed a recurrent neural network architecture that leverages cycle consistency and human MoCap data for computer animation motion retargeting. 
However, their method is limited to cases where the kinematic structure only differs in terms of bone lengths and proportions (i.e. same topology). 
To enable motion retargeting of different skeleton topologies, \cite{yan2023imitationnet} proposed a human-to-robot unsupervised motion retargeting method that constructs a shared latent space through adaptive contrastive learning and incorporates a consistency term. 
Our approach, however, examines fully supervised learning with a large and automatically collected paired dataset.

To automatically collect a paired dataset, Choi et al. \cite{choi2021self} proposed a $\langle \text{robot, human}\rangle$ pose pairing method that leverages the robot joint's movement information. 
Their method consists of randomly sampling robot poses from joint angles, more specifically from beyond the min-max range of the joint angles. 
Then, a nonparametric optimization method \cite{choi2020nonparametric} in the latent space is used to obtain feasible robot poses.
This nonparametric optimization method involves combining nonparametric regression and deep latent variable modeling techniques. 
The feasible robot pose is obtained by feeding the randomly sampled pose data to the domain-specific encoder and then decoding the largest feasible pose found by employing locally weighted regression on the latent space. 
After that, joint angles are transformed into a human pose using forward kinematics. 
Their method offers an advantage in terms of pose diversity by not limiting it to human MoCap data. 
This is achieved by avoiding the use of human MoCap data poses and instead employing randomly sampled robot poses when generating new pairs of poses.
However, their method relies on K-nearest neighbors search at the inference stage which makes it computationally inefficient \cite{choi2020nonparametric}. 
Also, it still inevitably requires paired data samples for the nonparametric optimization collected by using expensive IK solvers. 
Another crucial drawback is that randomly sampling from robot joints can result in infeasible or extreme human poses (i.e. noisy poses) which lower the quality of the obtained dataset (Fig. \ref{fig:1_Extreme_poses_example}-(b)).

Our method differs from these works because we enable fully supervised motion retargeting through the efficient collection of a large $\langle \text{robot, human}\rangle$ paired dataset. 
When collecting our data, a requirement is finding the hyperparameters of our system such as human-to-robot size scale factor. 
Basically, this can be designed by engineers, or be optimized with a small number of $\langle \text{robot, human}\rangle$ pose pairs.


\section{METHODOLOGY}
\begin{figure*}[h]
    \includegraphics[width=0.95\textwidth]{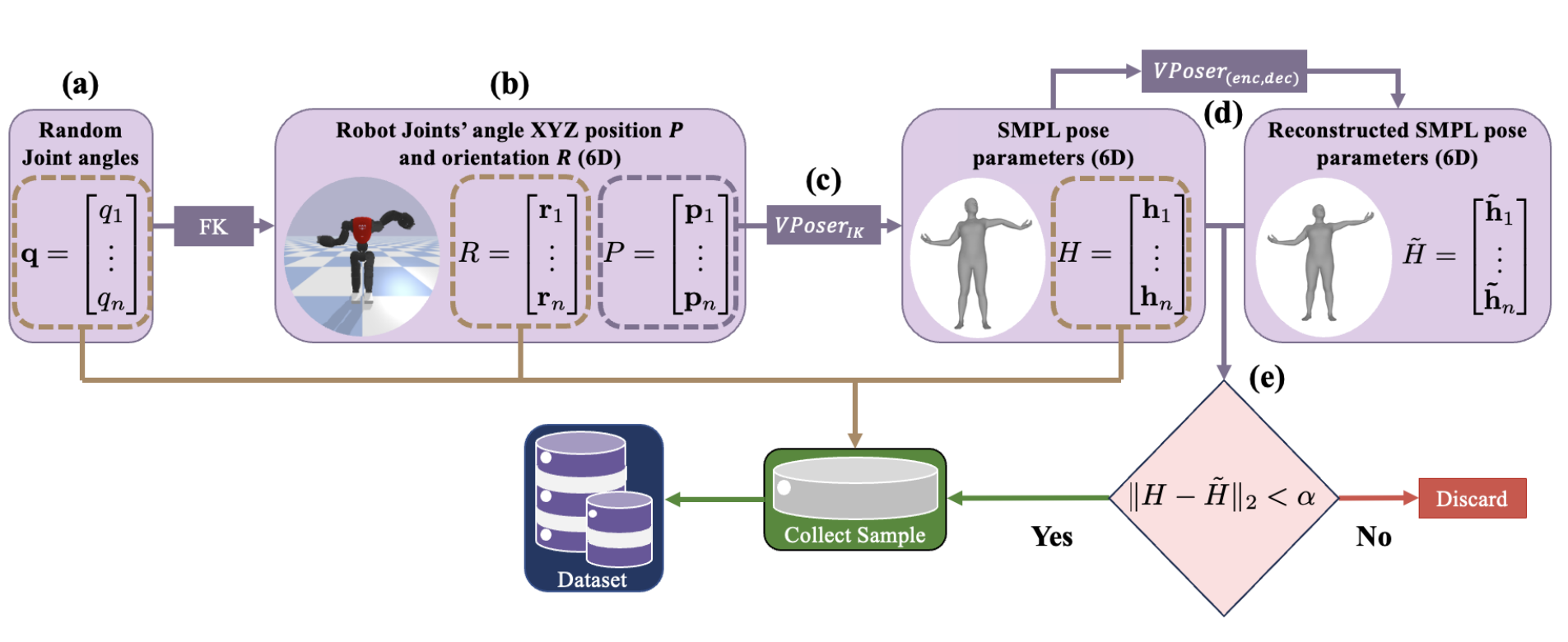} 
    \centering
    \caption{Overview of our data sampling process and extreme filtering component. (a) A robot pose $\mathbf{q}$ is sampled from the valid min-max joint angle range, (b) converted to the XYZ position $P$ and (6D) orientation $R$ by using forward kinematics. (c) The inverse kinematics solver $VPoser_{IK}$ uses $P$ to obtain (6D) SMPL human pose parameters $H$. Then, (d) by passing $H$ to the encoder and decoder of VPoser, its denoised version $\tilde{H}$ is obtained. Finally, (e) we measure the mean square error (MSE) between $H$ and $\tilde{H}$ and detect noisy poses. We discard the data if the MSE error is larger than a certain threshold.}
    \label{fig:Data_pairing}
    \vspace{-3mm}
\end{figure*}

\subsection{Overview}

Let $M= \begin{bmatrix} \mathbf{q}_1 & \ldots & \mathbf{q}_T \end{bmatrix} \in \mathbb{R}^{T \times n}$ denote an upper body robot motion, which is a sequence of joint angle vectors $\mathbf{q}_t$. 
Here, $\mathbf{q}_t = \begin{bmatrix} q_{t,1} & \ldots & q_{t,n} \end{bmatrix}^\top \in \mathbb{R}^n$ is an $n$ dimensional vector, where $n$ denotes the number of joints. 
We describe the orientation of robot links using a 6D representation (i.e. a continuous representation of 3D rotations) \cite{zhou2019continuity} as $R_t \in \mathbb{R}^{6 \times m}$, where $m$ is the number of robot links.
We employ SMPL\cite{SMPL:2015} pose parameters to represent a human pose, denoted as $H_t \in \mathbb{R}^{6 \times k}$, where $k$ is the number of joints in the SMPL body model.
While it is conventional to represent SMPL pose parameters with the axis-angle representation, we convert them into the 6D representation to ensure consistency with the representation used to describe the robot pose 3D orientation.

We present an end-to-end supervised learning framework with neural networks that learn a mapping from $H_t$ to $\mathbf{q}_t$. 
To accomplish this task with supervised learning, it is inevitable to collect a paired $\langle \text{robot, human}\rangle$ pose training dataset. Such datasets are usually obtained by solving inverse kinematics from human MoCap data to robot motion data \cite{choi2020nonparametric}, which is computationally expensive. 
Yet, we bypass the need to use human MoCap data as direct input and expensive inverse kinematics solvers during paired data collection. 
Our solution lies in an effective paired data collection method that deviates from traditional methods and obtains $H_t$ from $\mathbf{q}_t$. 
Our method allows the cost-efficient collection of large amounts of diverse and high-quality human poses (i.e. remove infeasible and extreme human poses), which we achieve by exploiting a human body prior (VPoser) \cite{SMPL-X:2019}.

In addition to our data pairing method, we also propose a two-stage network that maps $H_t$ to $\mathbf{q}_t$ by leveraging a continuous representation of 3D orientations (i.e. the 6D representation) \cite{zhou2019continuity}. 
Our two-stage network consists of a pre-network $\mathcal{F}_{pre}$ and a post-network $\mathcal{F}_{post}$. 
In the first stage, $\mathcal{F}_{pre}$ learns a mapping from $H_t$ to $R_t$ (Eq. \eqref{eq:first}), and in the second stage, $\mathcal{F}_{post}$ learns a mapping from the outputs of $\mathcal{F}_{pre}$ to $\mathbf{q}_t$ (Eq. \eqref{eq:second}).
At the real-world inference stage, we use a regression-based human mesh recovery network \cite{pymafx2023} to obtain $H_t$ from images $I_t$ of a video $\mathbf{V} = \{ I_1, \cdots, I_T \}$, which is a sequence of full human body images $I_t \in \mathbb{R}^{w \times h \times 3}$. 
We denote this process as $H_t = \mathcal{F}_{smpl}(I_t)$.

\begin{align}
R_t & = \mathcal{F}_{pre}( \mathcal{F}_{smpl}(I_t) )  \label{eq:first} \\
\mathbf{q}_t & = \mathcal{F}_{post}(R_t)  \label{eq:second}
\end{align}

\subsection{Data Pairing}

We introduce a new $\langle \text{robot, human}\rangle$ pose data pairing method for data-driven motion retargeting frameworks. 
Inspired by the concept of harnessing the robot joints' angles information for paired data collection \cite{choi2021self}, our pairing method deviates from the traditional methods that obtain pairs from a human pose to a robot pose \cite{choi2020nonparametric}. 
Our reverse process (i.e. obtaining $H_i$ from $\mathbf{q}_i$)  does not require any human motion capture data, thus the quality of $\mathbf{q}_i$ does not depend on the variety and size of the available MoCap data. 
Unlike \cite{choi2021self} which necessarily requires glue data (i.e. paired data generated by classic optimization-based methods), our method obtains feasible and natural human poses with no reliance on expensive IK solvers. 
We achieve this by leveraging a Multi-Person Linear model (SMPL) \cite{SMPL:2015} to represent human body poses and a Variational Human Pose Prior (VPoser) \cite{SMPL-X:2019} as both an efficient IK solver and a noisy poses detector. 

\begin{figure*}[h!tbp] 
    \centering    \includegraphics[width=0.82\textwidth]{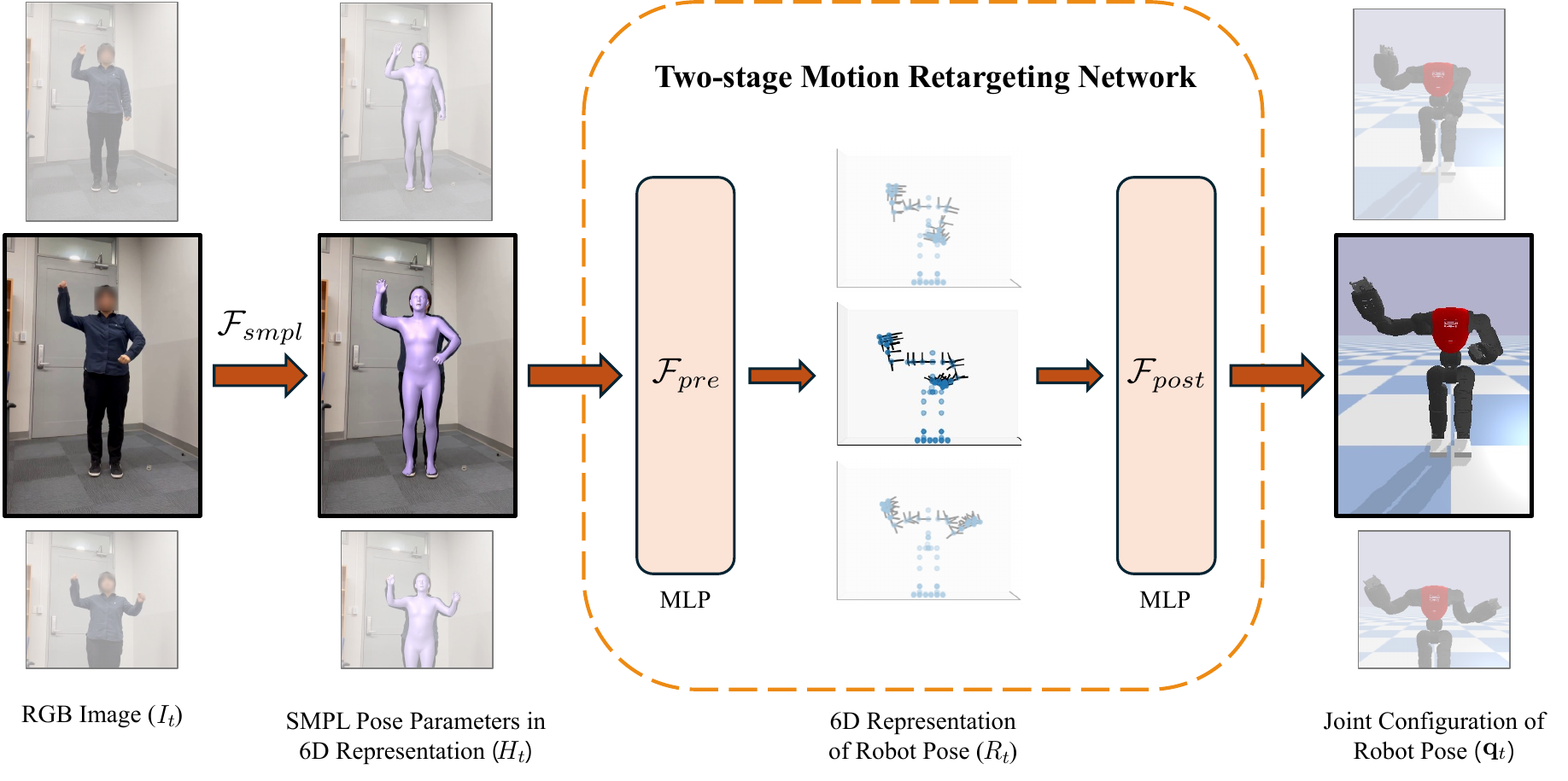}    
    \caption{The proposed supervised two-stage motion retargeting pipeline. An RGB image $I_t$ of a human pose from a video is converted to SMPL pose parameters using a mesh recovery network. These, (6D) SMPL pose parameters $H_t$ are input to the pre network $\mathcal{F}_{pre}$ which converts them to the corresponding robot pose (6D) orientation $R_t$. Then, $R_t$ is input to the post network $\mathcal{F}_{post}$ which maps it to the corresponding robot joint angles $\mathbf{q}_t$.}
    \label{fig:motion_retargeting_pipeline}
    \vspace{-5mm}
\end{figure*}

SMPL \cite{SMPL:2015} is a 3D mesh model of the human body, trained from thousands of 3D body scans and based on skinning and blend shapes. 
SMPL enables the generation of 3D human poses by utilizing pose parameters, which describe a human pose using the axis-angle representation of human body joints. 
An advantage of employing the SMPL model is that we can efficiently solve IK and obtain pose parameters by using VPoser's solver \cite{SMPL-X:2019}. 
VPoser is a variational autoencoder trained from a large dataset \cite{mahmood2019amass} of human poses represented as SMPL parameters which allows it to learn a probability distribution of human body poses. 
The efficiency of VPoser is primarily due to its optimization of a low-dimensional latent vector rather than the entire SMPL pose parameter set. 
This efficiency can be further increased with parallel computing with GPUs. 
Another significant advantage of VPoser is its capability to handle implausible poses. 
We achieve this through the reconstruction of poses, since we found out that VPoser can be used as a denoising autoencoder. 
If an infeasible and noisy human pose passes the encoder and decoder of VPoser, the obtained pose becomes the denoised version of the input pose. 
Thus, we use these reconstructed poses to assess errors and guarantee adherence with valid 3D human poses.  

\begin{algorithm}[t]
\SetAlgoLined
\KwIn{Number of samples \(N\), \(
[q_{\min}^{j}, q_{\max}
^{j}]_{\{j=1,...,n\}} \) } 

\KwOut{Dataset \( D \) =  $\{\mathbf{q}_i, R_i, H_i\}_{\{i=1,..., N\}}$, Reconstruction errors array \( \Phi \)}

Initialize $\text{VPoser}_{(IK,enc,dec)}$ 

Initialize \( D \) and \( \Phi \) with length \( N \)\
    
\For{i $=1, N$}{

Initialize \(\mathbf{q}_i\) as an empty vector\
      
\For{\( j = 1 \) \KwTo \( n \)}{

Sample $q_j$ randomly $q^{j}_{\min} \leq \ q_j \leq q^{j}_{\max}$

Append \(q_j\) to \(\mathbf{q}_i\)\
}

Solve Forward Kinematics $R_i, P_i = FK(\mathbf{q}_i)$

Compute human pose $H_i = \text{VPoser}_{IK}(P_i)$

Encode human pose $z_i = \text{VPoser}_{enc}(H_i)$

Compute mean latent pose $\bar{z}_i = \text{mean}(z_i)$

Reconstruct human pose $\tilde{H}_i = \text{VPoser}_{dec}(\bar{z}_i)$

Calculate reconstruction error: $ \phi_i = \| H_i - \tilde{H}_i \|_F $
    
Store \( \phi_i \) in \(  \Phi \)\
    
Append $\langle \mathbf{q}_i, R_i, H_i \rangle$ to $D$ 
}
\Return $D$, $\Phi$ 
\caption{Data Pairing Algorithm}
\label{alg:data_pairing}
\end{algorithm}

The applicability of our method extends to any humanoid robots, with several hyperparameters as requirements. 
The hyperparameters are mainly related to the size scale factor, and they compensate for the mismatch between coordinates due to the different proportions and height differences between humans and robots. 
This scale factor can be manually designed by engineers, considering the link lengths of humans and robots. 
But if one is determined to optimize the hyperparameters, one can find the sub-optimal ones with a small set of $\langle \text{robot, human}\rangle$ pose pairs obtained by using a traditional IK solver. 
In our experiment, we manually designed the hyperparameters for the Reachy and Coman robots, and found sub-optimal ones for the Nao robot. 
To find the sub-optimal ones, we define our data pairing method as a function $g_\theta : \mathbf{q} \to H$, where the scaling factor $\theta$ is a set of hyperparameters. 
Then, we randomly sample $\theta$ from a valid range and measure the error between $g_\theta(\mathbf{q})$ and $H$. 
Among the randomly sampled $\theta$, we choose the one which makes the minimum error. 

Once this robot description information is available, our process (Alg. \ref{alg:data_pairing}) to obtain a paired dataset  \( D \) starts with acquiring robot poses (Fig. \ref{fig:Data_pairing}(a)) for which we randomly sample $\mathbf{q}_i$ from the valid range of robot joint angles.
Then, we solve Forward Kinematics (Fig. \ref{fig:Data_pairing}(b)) and input the joints' XYZ position to VPoser's solver which returns $H_i$ (Fig. \ref{fig:Data_pairing}(c)). 
When sampling poses from random robot joint angles, a major issue is that $H_i$ can result in a pose that is infeasible and extreme (i.e. noisy poses). 
We detect these noisy poses by computing the deviation of $H_i$ to denoised human poses (Fig. \ref{fig:Data_pairing}(e)). 
For this, we encode $H_i$ to the low dimensional latent space of VPoser (Fig. \ref{fig:Data_pairing}(d)), compute the mean latent vector, and decode it to obtain the reconstructed pose $\tilde{H}_i$. 
Then, we compute the reconstruction error $\phi$ between $H_i$ and $\tilde{H}_i$ as follows:%
\begin{align}
    \phi = \text{MSE}(H_i, \tilde{H}_i) = \frac{1}{k} \sum_{l=1}^{k} (H_{i,l} - \tilde{H}_{i,l})^2
\end{align}%
To detect noisy poses, we first obtain a normal distribution of the reconstruction errors and set a threshold which we empirically found to be one standard deviation above the mean. 
From the dataset \( D \), we remove the pose data $\langle \mathbf{q}_i, R_i, H_i \rangle$ whose reconstruction error is larger than the threshold. 
To sample a batch from the remaining poses in the training stage, a Bernoulli distribution with probability $P_r$ is used, where $P_r$ is obtained from a function that is negatively and linearly correlated to the reconstruction error $\phi$.

\subsection{Two-Stage Motion Retargeting Network}

We propose a supervised learning model to learn a mapping from $H_t$ to $\mathbf{q}_t$ for data-driven motion retargeting. 
Considering that neural networks tend to learn better from a continuous representation of orientation, we propose a two-stage network that exploits the 6D representation of orientation \cite{zhou2019continuity}.
We divide the process into two distinct mappings as shown in Figure \ref{fig:motion_retargeting_pipeline}.
These two separate mappings are learned by two distinct networks, pre-network $\mathcal{F}_{pre}$ and post-network $\mathcal{F}_{post}$, respectively. 
First, $\mathcal{F}_{pre}$ learns a mapping from $H_t$ to $R_t$ (i.e. poses in different domains) where the orientation of both $H_t$ and $R_t$ are described using the 6D representation \cite{zhou2019continuity}. 
In the second stage, $\mathcal{F}_{post}$ learns a mapping from $R_t$ to $\mathbf{q}_t$. The purpose of $\mathcal{F}_{post}$ is to solve an IK problem between poses of the same domain (i.e. robot poses). 

Both $\mathcal{F}_{pre}$ and $\mathcal{F}_{post}$ are fully connected networks with one hidden layer. 
Our model employs a simple architecture to underscore the advantages in computational efficiency and accuracy that neural networks can achieve when a high-quality dataset is available. 
Designing more complex architectures would be our future work, but simpler architectures can be efficient for real-time motion retargeting.

For the training of the whole network, we compute and combine the losses from the two networks (i.e. $\mathcal{F}_{pre}$ and $\mathcal{F}_{post}$). 
We define the loss for the pre-network $\mathcal{L}_{pre}$ as the summation of the mean squared error between the predicted $\hat{R}_t$ and the ground truth $R_t$. 
Then for the learning of $\mathcal{F}_{post}$, we use a teacher-student learning technique for which we compute the post-network loss $\mathcal{L}_{post}$, employing two predicted joint angle vectors: a teacher joint angle vector $\hat{\mathbf{q}}_t$ from the ground truth $R_t$, and a student joint angle vector $\hat{\mathbf{q}}'_t$ from the predicted $\hat{R}_t$. 
Then, $\mathcal{L}_{post}$ is defined as the summation of the mean squared error between $\hat{\mathbf{q}}_t$ and $\mathbf{q}_t$ and mean squared error between $\hat{\mathbf{q}}'_t$ and $\mathbf{q}_t$. 
Lastly, the total loss $\mathcal{L}_{total}$ is defined as a summation of $\mathcal{L}_{pre}$ and $\mathcal{L}_{post}$.

\begin{align}
    \mathcal{L}_{pre} & =  MSE(R_t, \hat{R}_t) \\
    \mathcal{L}_{post} & = MSE(\mathbf{q}_t, \hat{\mathbf{q}}_t) + MSE(\mathbf{q}_t, \hat{\mathbf{q}}'_t) \\
    \mathcal{L}_{total} & = \mathcal{L}_{pre} + \mathcal{L}_{post}
\end{align}

\section{EXPERIMENTS}

\begin{figure*}[h!tbp]
    \includegraphics[width=0.95\textwidth]{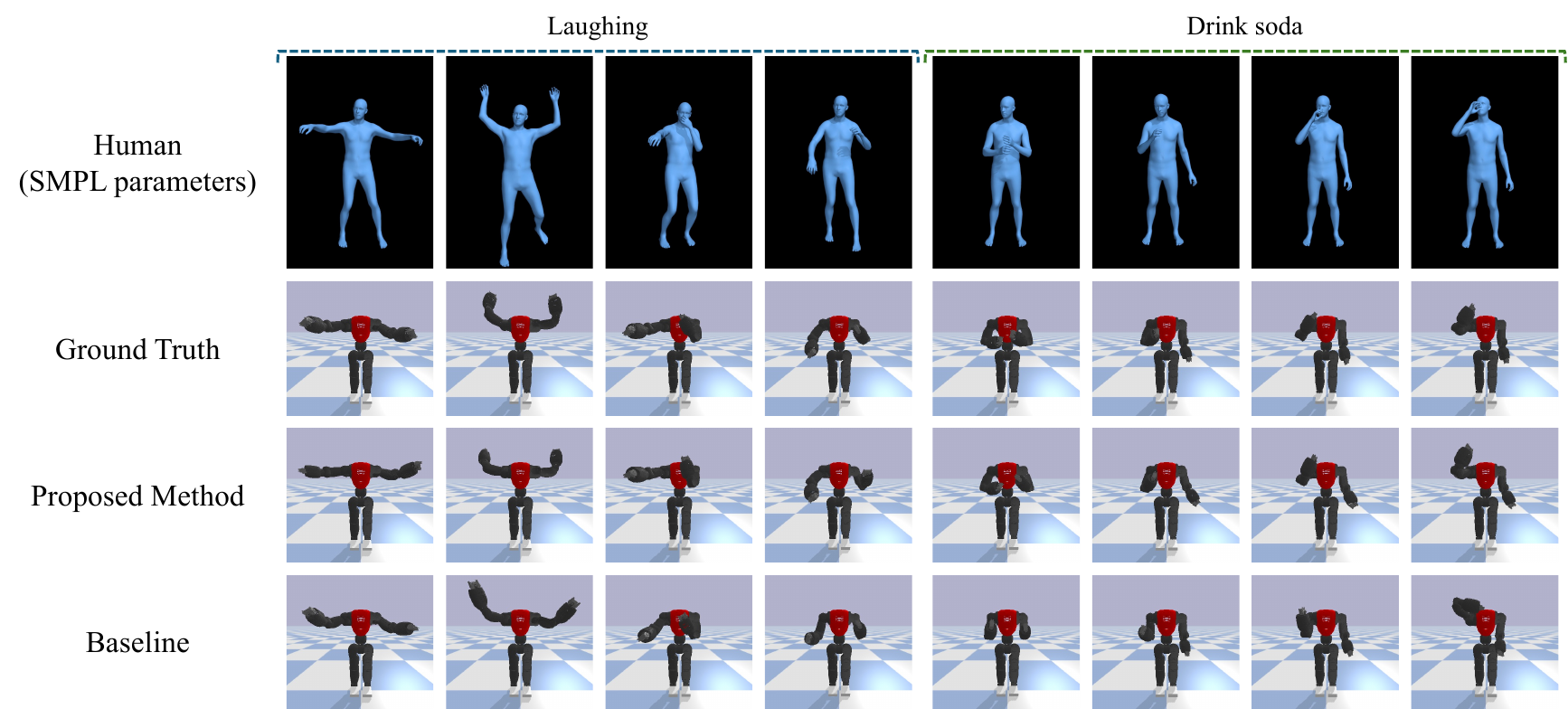}
    \caption{Visualization showcasing the performance of our method in determining XYZ positions of links relative to the baseline on evaluation set poses.
    }
    \label{fig:key_poses}
\end{figure*}

In this section, we demonstrate the capabilities of our paired data collection method to generate high-quality $\langle \text{robot, human}\rangle$ pairs and the effectiveness of our proposed two-stage network for supervised motion retargeting. 
 To fulfill this goal, we design three experiments: 
 (1) a comparison between the motion retargeting performance of our fully supervised learning-based method and the unsupervised learning-based method introduced in \cite{choi2021self} which we chose as our baseline (Section \ref{sec:comparing_with_baseline}), 
 (2) an ablation study to analyze the effects of our extreme pose filtering component and our two-stage network architecture (Section \ref{sec: data pairing ablation study}) and 
 (3) a real-time motion retargeting experiment in a real-world setting (Section \ref{sec: real-world motion retargeting}).  
 In experiment 2, we also examine the scalability of our extreme pose filtering component to three robots (i.e. Reachy, Coman and Nao) with different kinematics and topology (i.e, number of joints and sizes). 
For Coman and Nao, all experiments were conducted exclusively within a simulation environment. For Reachy, experiments were carried out both in a simulated environment and in a real-world setting (e.g., experiment 3).

\subsection{Dataset}
For our experiments, we generated several datasets. 
First, for the training of our two-stage neural network, we collected two million $\langle \text {robot, human}\rangle$ pose pairs employing our robot-to-human pose paired data collection method. 
We obtained the data pairs for each one of the robots (i.e. Reachy, Coman and Nao) using each robot's description information from the unified robot description format (URDF). 
With an RTX A6000 GPU, collecting two million of pairs took approximately one and a half day for each robot. 
We also generated $\langle \text{robot, human}\rangle$ paired evaluation datasets per robot using an optimization-based \cite{choi2019towards} method. 
These evaluation datasets are not fed into the model during training, and are used as a ground truth for a fair quantitative evaluation. 
For this dataset, we selected motions from the AMASS dataset \cite{mahmood2019amass} that have higher variance in upper body movement, with minimal or no movement of the lower body. 
This evaluation set is composed of the motion of 7 different subjects and a total of 11 different motions: \textit{punch, drink soda, laugh, boxing, wash windows, direct traffic, hand signals, basketball signals, superhero, panda, and vignettes}. 
For training our baseline \cite{choi2021self}, we collected the dataset by following the details in their work and also used additional glue data to learn the shared latent space for the nonparametric optimization algorithm they use.

\subsection{Evaluation Metrics}
For a more comprehensive evaluation of our method, we measure the performance using two different evaluation metrics in all our quantitative experiments — one for assessing the predicted joint angles and the other for evaluating the link XYZ position of the pose after retargeting. 
For the joint angle error, we compute the absolute difference while considering the periodicity of 2$\pi$, between the predictions of our model or the baseline and the ground truth robot joint angles.
For the link distance error, we solve FK to get the XYZ position of the links, select the key links our robot moves, and calculate the Euclidean distance.

\subsection{Motion Retargeting Network Training Setup}

Our motion retargeting network consists of two networks. Each network is a multi-layer perceptron of one hidden layer with 512 neurons. 
The two-stage network is trained using the GELU \cite{hendrycks2016gaussian} activation function, a learning rate of 1e-4, and a batch size of 2048. 
For the learning, we use the MSE loss which is optimized using the Adam optimizer \cite{kingma2014adam} with a weight decay of 1e-6. 
We chose the best model parameters by using a validation set selected from the evaluation set. 
More specifically, since our ground truth dataset contains 11 motions, we selected 2 motions for the validation set and 9 motions for the test set. 
To ensure the validation set had varied robot joint angles we selected those motions that had the higher variance among the 11 motions. 

\begin{table}[t]
    \centering
    \caption{A comparison of the performance of our proposed method and the baseline onto Coman.
    }
    \label{tab:basline_comparison}
    \begin{tabular}{lcc}
        \toprule
        
        Method & Joint Diff. (rad) & Link Distance (cm) \\
        \midrule
        Baseline & \textbf{0.281} & 5.956 \\
        Ours & 0.295 & \textbf{4.913} \\
        
        \bottomrule
    \end{tabular}
    \vspace{-4mm}
\end{table}

\subsection{Comparison to the Unsupervised Learning Method
} \label{sec:comparing_with_baseline}

Our aim is to collect a $\langle \text{robot, human}\rangle$ pose paired dataset to enable the training of a model with a fully supervised learning method since such methods tend to perform better than unsupervised learning methods when high-quality labeled data is available. 
In this section, we evaluate the performance of our fully supervised motion retargeting method when retargeting motion onto Coman, compared to an unsupervised motion retargeting baseline \cite{choi2021self}.
The results (Tab. \ref{tab:basline_comparison}), show that while the baseline is better at joint angle prediction, our proposed supervised method yields better results regarding link XYZ position (Fig. \ref{fig:key_poses}). 
Another important aspect to consider is the computational efficiency. 
In inference stage, the baseline method employs the K-nearnest neighbors algorithm to obtain the robot pose that is closest to the input human pose in the shared latent space. 
However, this approach renders the method computationally inefficient since it needs to find the nearest one from a large number of candidates. 
On the other hand, our method is based on a deep neural network, which computation is fast at the inference stage.

\begin{table}[t]
    \centering
    \caption{
    Ablation study of our extreme pose filtering component and Two-Stage Network Architecture.
    }
    \label{tab:ablation_study}
    \begin{tabular}{lcccc}
        \toprule
        
        Robot & Error & Proposed & (- Pose Filter) & (- Two-stage) \\
        \midrule
        Reachy & Joint (rad) & \textbf{0.253} & 0.257 & 0.263 \\
               & Link (cm) & \textbf{8.678} & 8.698 & 9.519 \\
        \midrule
        Coman & Joint (rad) & \textbf{0.295} & 0.297 & 0.314 \\
               & Link (cm) & \textbf{4.913} & 5.008 & 5.148 \\
        \midrule
        Nao & Joint (rad) & \textbf{0.389} & 0.413 & 0.506 \\
               & Link (cm) & \textbf{3.887} & 3.956 & 4.895 \\
        \bottomrule
    \end{tabular}
    \vspace{-2mm}
\end{table}

\subsection{Ablation Study: Pose Filtering and Network Architecture} \label{sec: data pairing ablation study}

\textcolor{black}{
In this section, we analyze the effects of (1) the extreme pose filtering component in our data pairing method, and (2) our two-stage network design.
Table \ref{tab:ablation_study} shows the results of our ablation study, where \textit{Proposed} is the method using both extreme pose filtering and two-stage network, \textit{(- Pose Filter)} is the result of training the two-stage network on data without pose filter, and \textit{(- Two-stage)} is the result of training the one-stage network on data with pose filter.}

In the ablation study of the pose filter, we trained our two-stage network for Reachy, Coman and Nao with and without our proposed extreme pose filtering method. 
For a fair comparison between the two methods, it is necessary to train the model with the same hyperparameters such as batch size and number of iterations. 
However, if we set the batch size of the two cases to be the same, the model with extreme pose filtering will be trained with a smaller amount of data per epoch, since the size of the whole training data becomes smaller after the filtering.
To compensate for this issue, when training our method with extreme pose filtering, we increased the number of epochs and batch size to be larger than that of the method without filtering, so that a number of iterations can be similar for both approaches.
The performance of our extreme pose filtering method achieved better results for both the joint and link error compared to the method without filtering noisy poses.
The result shows that extreme pose filtering is much useful when the topology (i.e. number of joints) and the size of the robot is significantly different from the human. 
For instance, the performance gain in Nao, the small-sized humanoid robot with fewer joints, is larger than other robots.

\textcolor{black}{
To check whether the two-stage network is beneficial, we compare the proposed model with the one-stage model, which directly maps SMPL human pose parameters $H_t$ to the robot joint angles $\mathbf{q}_t$.
The results show that our two-stage network outperforms the one-stage network for both metrics across all robots, representing our two-stage network can learn the complex relationship between the human and robot poses better than the one-stage network.}



\subsection{Motion Retargeting in Real-world}
\label{sec: real-world motion retargeting}

\begin{figure}[]
    \includegraphics[width=\linewidth]{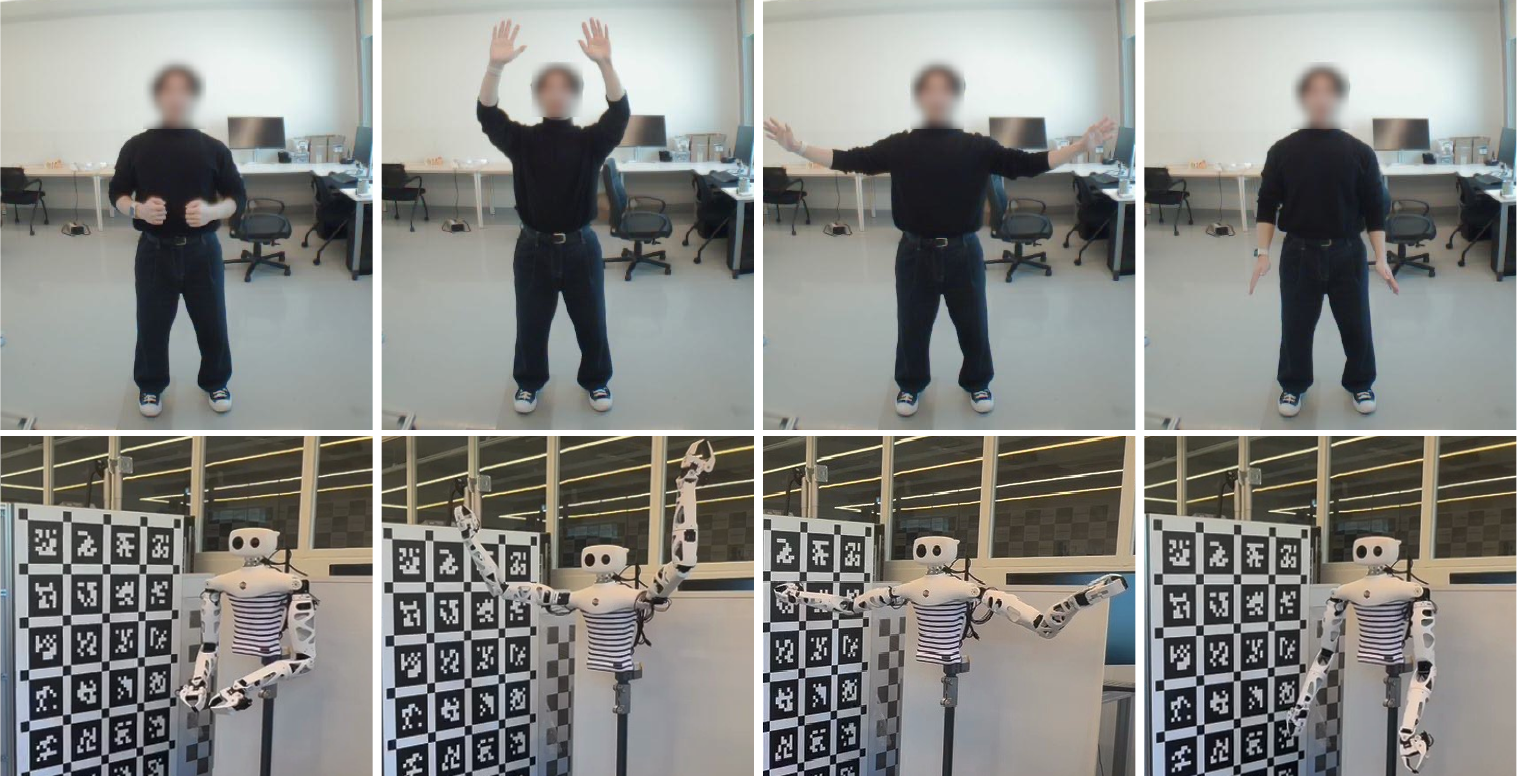}
    \caption{Real-time retargeting of human poses from RGB images onto Reachy in a real world environment.
    }
    \label{fig:rgb}
    \vspace{-5mm}
\end{figure}

Our proposed method can be easily integrated into real-world environments during the inference stage. This is facilitated by our representation of the human pose using SMPL pose parameters. Specifically, we leverage a human mesh recovery network \cite{pymafx2023} to obtain these parameters from an RGB camera. We conducted a real-time experiment (Fig. \ref{fig:rgb}) in a real-world environment to test our motion retargeting method on Reachy. Our goal was to qualitatively evaluate how effectively our method performs in practical settings.

        

\section{CONCLUSION}
\textcolor{black}{In this work, we propose a framework for fully supervised motion retargeting, which consists of a novel $\langle \text{robot, human}\rangle$ pose data pairing method as well as a two-stage neural network for retargeting. 
Our method converts random robot poses into human poses using VPoser, which can find SMPL human pose parameters corresponding to given XYZ joint positions of a robot. 
However, since a randomly sampled robot pose can result in an implausible or extreme human pose, our method filters it out based on the distance between its denoised version, since a result of passing a pose to the VPoser that can work as a denoising autoencoder. 
Then, we use collected $\langle \text{robot, human}\rangle$ pose data to train our two-stage network which exploits the advantage of 6D representation. Our experiments show that the proposed method outperforms the baseline, a method without extreme pose filtering, and a method with a one-stage network.
}
We also verified that our proposed method can be applied in a real-world environment.

\section*{ACKNOWLEDGEMENT}
This work was supported by the National Research Foundation of Korea (NRF) grant funded by the Korea government (MSIT)(RS-2023-00211416), by the Settlement Research Fund (1.220117.01) of UNIST (Ulsan National Institute of Science \& Technology), by Institute of Information \& communications Technology Planning \& Evaluation(IITP) grant funded by the Korea government(MSIT)(No.RS-2020-II201336, Artificial Intelligence graduate school support(UNIST)), and was partly supported by Innovative Human Resource Development for Local Intellectualization program through the Institute of Information \& Communications Technology Planning \& Evaluation(IITP) grant funded by the Korea government(MSIT)(IITP-2024-RS-2022-00156361, 25\%)
\bibliographystyle{IEEEtran}
\bibliography{citations}

\end{document}